\documentclass{article}
\usepackage[final]{colm2026_conference}
\usepackage[T1]{fontenc}

\usepackage{microtype}
\usepackage{hyperref}
\usepackage{url}
\usepackage{booktabs}
\usepackage{amsmath,amssymb}
\usepackage{graphicx}
\usepackage{multirow}
\usepackage{enumitem}
\usepackage{array}
\usepackage{tabularx}
\usepackage{pgfplots}
\pgfplotsset{compat=1.18}
\usepgfplotslibrary{fillbetween}

\usepackage{caption}
\usepackage{float}

\definecolor{darkblue}{rgb}{0, 0, 0.5}
\hypersetup{
  colorlinks=true,
  citecolor=darkblue,
  linkcolor=darkblue,
  urlcolor=darkblue,
  pdftitle={Logarithmic Scores, Power-Law Discoveries: Disentangling Measurement from Coverage in Agent-Based Evaluation},
  pdfauthor={HyunJoon Jung and William Na}
}

\graphicspath{{figures/}}

\title{Logarithmic Scores, Power-Law Discoveries:\\Disentangling Measurement from Coverage in\\Agent-Based Evaluation}

\author{%
HyunJoon Jung\thanks{Corresponding author: \texttt{hjung@mphora.ai}}\\
MPhora\\
\texttt{hjung@mphora.ai}
\And
William Na\\
MPhora\\
\texttt{won@mphora.ai}
}

\begin{document}

\maketitle

\vspace{-0.5em}
\begin{center}
\includegraphics[width=\columnwidth]{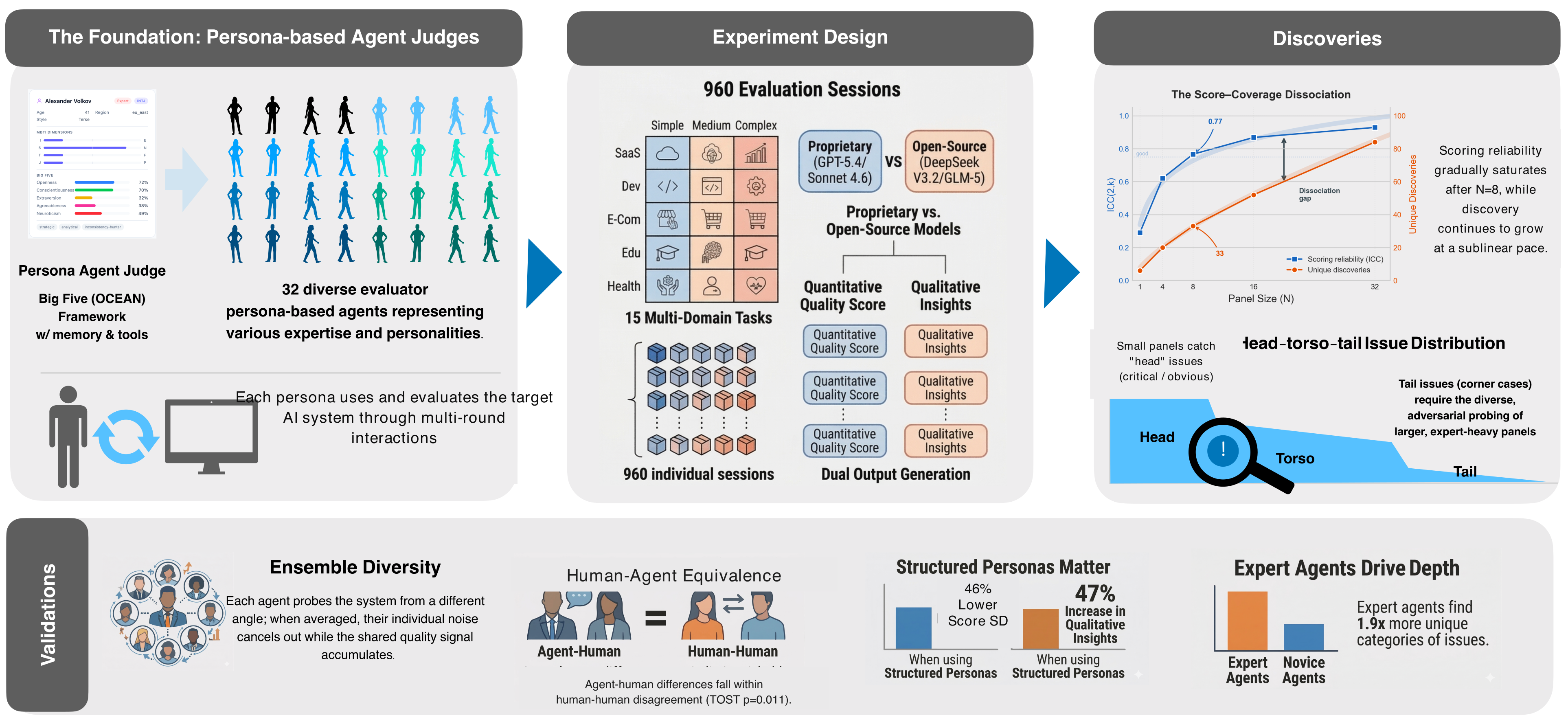}
\captionof{figure}{Overview. 32 persona-based agent judges evaluate conversational AI across 960 sessions. A task-level equivalence test finds that agent--human score differences fall within the range of human--human disagreement (90\% CI $[-0.038, 0.012]$). Scoring reliability improves logarithmically while issue discovery follows a sublinear power law ($b \approx 0.69$). A full-scale no-persona control reproduces both curve shapes; a separate controlled ablation shows that structured persona conditioning improves score consistency and issue yield.}
\label{fig:overview}
\end{center}
\vspace{-0.5em}


\begin{abstract}
LLM-based agent judges are an emerging approach to evaluating conversational AI, yet a fundamental uncertainty remains: \emph{can we trust agent judges to sufficiently cover and reliably score the failure modes of conversational AI systems?}
Through 960 sessions with two model pairs across 15 tasks, we show that persona-based agent judges produce score differences from human raters that fall within the range of human--human disagreement.
We then identify a \textbf{score--coverage dissociation}: scoring reliability improves logarithmically with panel size, while unique issue discoveries follow a sublinear power law---both exhibit diminishing returns, but scores saturate roughly twice as fast as discoveries.
We hypothesize this reflects a power law distribution of the finding space: critical issues are discovered first by small panels, while corner cases require progressively larger panels, analogous to species accumulation curves in ecology.
The mechanism traces to ensemble diversity, with expert judges acting as adversarial probes that push discovery into the tail of the finding distribution.
A full-scale control with 32 identical agents without persona conditioning reproduces both scaling shapes, indicating that they are general to stochastic multi-session evaluation; in a separate controlled ablation, structured persona conditioning reduces score SD by 46\% and produces 47\% more insights per session.
\end{abstract}


\section{Introduction}

Evaluating conversational AI now relies heavily on LLM-as-Judge approaches \citep{zheng2023judging, liu2023geval, kim2024prometheus}, which offer scalability but have well-documented biases in single-judge settings \citep{wang2024large}.
A natural response is to use \emph{multiple} evaluators---analogous to crowdsourced annotation, where aggregating multiple raters improves label reliability \citep{sheng2008get, snow2008cheap}.
But unlike human annotators labeling static examples, \textbf{agent judges} actively \emph{converse} with the system under test.
Each agent judge conducts a multi-turn interaction from a distinct persona---varying in expertise, communication style, and evaluation focus---and produces both a quality score and qualitative insights.

The standard assumption from crowdsourcing research \citep{sheng2008get, snow2008cheap} is that more raters yield more reliable consensus.
We show that this assumption is \textbf{incomplete} for agent-based evaluation.
In crowdsourcing, all annotators label the \emph{same} item; in agent-based evaluation, each judge generates a \emph{different interaction}, observing different facets of the same system.
This distinction changes what ``adding more raters'' can achieve.
The design is analogous to a multi-round interview process: aggregating independent interviews can stabilize an overall assessment while each additional interviewer also uncovers different evidence.

Evaluating conversational AI well requires both \emph{reliable scoring}---can we trust the quality rating?---and \emph{broad coverage}---have we observed enough nondeterministic behavior to understand where the system fails?
These seem naturally aligned: more judges should improve both.
To test this, we conducted a large-scale study using 32 persona-based agent judges---each with distinct expertise, Big Five personality profile \citep{costa1992neo}, and interaction style---across 15 tasks spanning 5 domains and 3 complexity levels (480 sessions per model pair, 960 total).

A Turing-style comparison with 50 recruited human raters finds that agent--human score differences fall within the range of human--human disagreement (TOST 90\% CI $[-0.038, 0.012]$, $p = 0.011$ at an equivalence margin of $0.05$).
With this comparison in place, the scaling analysis reveals a \textbf{score--coverage dissociation}: scoring reliability improves logarithmically with panel size, while unique issue discoveries follow a sublinear power law---both exhibit diminishing returns, but scores saturate roughly twice as fast as discoveries.
An \textbf{expertise effect} sharpens this picture: expert agent judges score lower (Cohen's $|d| = 0.24$) by posing harder queries that expose quality limitations invisible to simpler interactions.
Panel reliability, meanwhile, arises not from agents agreeing with each other, but from agents \emph{probing different aspects} of the same system---their diverse errors cancel out while the shared quality signal accumulates.
A full-scale control with identical agents without persona conditioning reproduces both scaling shapes, while a separate controlled ablation attributes lower score dispersion, more insights per session, and a moderated expertise effect to structured persona conditioning.

The contributions of this study are as follows: (1)~a task-level equivalence analysis (50 human raters recruited, 43 retained after removing incomplete or duplicate submissions, 86 sessions) finds that agent--human score differences fall within the range of human--human disagreement; (2)~we identify the score--coverage dissociation---scoring reliability improves logarithmically while issue discovery follows a sublinear power law---and show it holds across two model pairs; (3)~a variance decomposition reveals the ensemble diversity mechanism behind panel reliability, and we show that expert judges function as adversarial probes, discovering $1.9\times$ more issue categories; and (4)~a 480-session no-persona control shows that both scaling shapes are general to stochastic multi-session evaluation, while a separate controlled ablation shows the level improvements from structured persona conditioning.


\section{Related Work}

\paragraph{Multi-rater aggregation.}
\citet{dawid1979maximum} introduced the EM framework for estimating true labels from noisy annotators.
\citet{snow2008cheap} showed that 4--5 non-expert crowdworkers can match expert quality, and \citet{sheng2008get} demonstrated that selectively acquiring additional labels improves data quality.
These studies established that more raters generally help---but all assume raters label the \emph{same} static item.
We extend this question to agent-based evaluation, where each rater generates a \emph{different interaction}, influencing how aggregation works.

\paragraph{Beyond scoring: why coverage matters.}
Most evaluation frameworks focus on quality scores, but scores miss \emph{what issues the system has}.
In usability testing, \citet{nielsen1993mathematical} showed different users discover different problems; \citet{faulkner2003beyond} and \citet{schmettow2012sample} agreed: \emph{coverage} follows a different curve than \emph{reliability}.
For generative AI, this gap matters more---nondeterministic outputs mean a single interaction cannot capture the full range of behavior.
In ecology, species accumulation curves \citep{colwell1994estimating} formalize a similar phenomenon: common species are found first, rare species require more sampling.
To our knowledge, no prior work has applied this framing to agent-based evaluation or formally disentangled scoring and coverage scaling within the same framework.

\paragraph{Diversity-driven red teaming.}
Automated red teaming also uses diverse interactions to broaden behavioral coverage.
Rainbow Teaming formulates adversarial prompt generation as quality-diversity search \citep{samvelyan2024rainbow}, while Anthropic surveys complementary expert, community, and automated approaches \citep{anthropic2024redteam}.
Our setting is broader than safety-specific red teaming, but shares the premise that independent probes expose different failure modes.

\paragraph{From LLM-as-Judge to Agent-as-Judge.}
\citet{zheng2023judging} popularized the LLM-as-Judge framing; follow-up work tackled biases \citep{wang2024large, wang2025judgingbias}, multi-agent debate \citep{chan2024chateval}, and pairwise discussion \citep{li2024prd}.
The Agent-as-a-Judge paradigm \citep{zhuge2025agent} (ICML 2025) and IntellAgent \citep{levi2025intellagent} extended this to interactive evaluation; \citet{verga2024poll} showed a panel of smaller LLMs outperforms a single GPT-4 judge; \citet{tseng2025expertannotators} found discussion on the \emph{same} item yields marginal gains---our agents independently generate \emph{different} interactions.
Surveys \citep{yu2025survey, you2026agentjudge} identify panel composition and scaling as open problems we address directly.

\paragraph{Persona-based agents.}
\citet{park2023generative} demonstrated realistic persona agents; \citet{serapiogarcía2023personality} and \citet{jiang2024evaluating} showed that LLMs can be prompted to express Big Five profiles.
\citet{shu2024psychometric} found that minor prompt perturbations can make LLM responses to psychometric instruments unreliable, and \citet{zou2025selfreport} found self-report personality correlates weakly with interaction behavior.
We therefore do not infer persona fidelity from model self-report; we validate evaluation behavior directly.
\citet{pal2025personas} likewise show that character-consistent dialogue depends on how persona information is grounded, motivating our persistent state and diary-based anchors.
Most recently, \citet{lu2025uxagent} deployed persona-aligned LLM agents to simulate usability testing of web designs.
Our focus is on the \emph{statistical properties} of agent panels---how scores and discoveries scale with panel size.

\paragraph{Reliability theory and ensembles.}
The Spearman-Brown formula \citep{spearman1910correlation} predicts reliability growth assuming tau-equivalent raters.
\citet{shrout1979intraclass} formalized ICC for various designs.
In machine learning, \citet{krogh1994neural} showed ensemble accuracy depends on member \emph{diversity}, not just individual accuracy---a decomposition later unified by \citet{wood2023unified}.
We draw an explicit analogy: agent-judge panel reliability arises from diverse perspectives rather than consistent agreement, just as ensemble prediction accuracy arises from diverse errors rather than individual accuracy.


\section{Method}

\subsection{Persona-Based Agent Judge Design}

Conversational AI systems behave differently depending on \emph{who} is interacting with them: an expert asks probing follow-ups that expose depth limitations, while a novice accepts surface-level answers that mask the same limitations.
A single generic judge---or a panel of identical judges---cannot capture this context dependence.
What is needed are \emph{contextual judges}: evaluators whose backgrounds, expertise, and communication styles shape the interaction itself, so that the evaluation reflects the range of real user experiences.

Persona-based design addresses this directly.
Rather than generating diversity through random sampling or temperature variation, we ground each agent judge in a structured persona that determines \emph{how} it interacts, \emph{what} it pays attention to, and \emph{how strictly} it evaluates.
We use the Big Five personality framework \citep{costa1992neo}---a widely used dimensional model of personality, and one that recent work shows LLMs can be prompted to express \citep{serapiogarcía2023personality, jiang2024evaluating}.
We chose the Big Five over alternatives (e.g., MBTI) because it is \emph{continuous} (enabling fine-grained differentiation), \emph{multidimensional} (enabling complementary trait profiles), and \emph{empirically grounded} in established psychometric research.
Big Five conditioning is what makes this diversity \emph{systematic}: different personality profiles lead agents to ask different questions, tolerate different failure modes, and weight different quality dimensions.
We deploy 32 agent judges with diverse profiles: 8 expert, 17 intermediate, and 7 novice, spanning 9 geographic regions and ages 22--68 (full pool in Appendix~\ref{app:personas}).

\paragraph{Panel coordination.}
A \emph{coordinator agent} analyzes each task's domain and complexity, then ranks the 32 agents by \emph{task fitness}---a composite of domain expertise match, personality suitability (e.g., high-Conscientiousness agents rank higher for complex multi-step tasks), and expertise level.
Panels of size $N$ select the top-$N$ from this ranking.
This task-aware selection ensures that even small panels contain contextually relevant judges, and smaller panels are strict subsets of larger ones---so observed differences between sizes reflect marginal contributions, not resampling variance.

\paragraph{Dual-role interaction.}
Each agent operates in two interleaved roles per session: (1)~\textbf{conversationalist}---conducting multi-turn interaction conditioned on persona, task context, conversation history, and internal emotional state; and (2)~\textbf{evaluator}---recording per-turn \emph{diary entries} invisible to the target.
Each diary entry captures: a response quality score, a satisfaction rationale explaining the score from the persona's perspective, specific issues or strengths (categorized as functionality, accuracy, helpfulness, clarity, or safety), and updated emotional state.
To illustrate, consider two agents receiving the same target response---a generic troubleshooting guide for an account access error:

\begin{quote}
\small
\textbf{Expert CTO diary} (turn 3, score: 0.4): \emph{``The response lists generic steps (clear cache, restart browser) but ignores the 403 status code I mentioned. No RBAC-level diagnosis offered.''} \textbf{Issue:} accuracy---ignores user-provided technical details.

\textbf{Novice Student diary} (turn 3, score: 0.8): \emph{``Clear instructions, easy to follow. Would be nice if it explained why this happens.''} \textbf{Issue:} helpfulness---lacks educational context.
\end{quote}

\noindent The same response produces a 0.4-point score gap and surfaces \emph{different} issues---a direct consequence of persona diversity.
The diary mechanism serves two purposes beyond recording scores.
First, it forces the agent to articulate \emph{why} it scored as it did, producing the qualitative insights that constitute our discovery data.
Second, the accumulated diary entries form a persona-consistent self-narrative that is fed back into the agent's context at each turn, reinforcing the assigned personality and reducing drift over longer conversations.

\paragraph{Emotional state dynamics.}
Real users do not evaluate each response in isolation: frustration compounds, trust erodes, and patience runs out over a conversation.
Each agent maintains a persistent emotional state across turns comprising five dimensions: \emph{trust}, \emph{frustration}, \emph{engagement}, \emph{patience}, and \emph{fatigue}.
Each dimension evolves after every turn based on the quality of the target's response, modulated by the agent's Big Five profile.
The emotional state is re-engaged at every turn through the personality-modulated update, acting as a recurring anchor to the assigned persona---a high-Neuroticism agent cannot drift toward calm behavior because its frustration state keeps being pushed upward.
Each agent also maintains a structured \emph{session memory} of all prior messages, target responses, diary entries, and emotional trajectory, enabling it to build on earlier observations and probe identified issues from different angles.

\subsection{Experimental Design}

The task set consists of 15 evaluation scenarios in a $5 \times 3$ factorial design: five domains crossed with three complexity levels.
The five domains---SaaS/IT, Developer, E-Commerce, Education, Healthcare---exercise distinct failure modes (factual precision, code accuracy, personalization, pedagogy, safety reasoning), each at three complexity levels (Simple/Medium/Complex: max 4/10/20 turns).

To test generalizability, we run the full experiment with two cross-family model pairs (judge $\neq$ target provider): a proprietary pair (GPT-5.4 target, Sonnet 4.6 judge) and an open-source pair (DeepSeek V3.2 target, GLM-5 judge, served via Friendli proxy).
Because recent evidence does not establish general, consistent model self-recognition \citep{davidson2024selfrecognition}, this separation is a conservative precaution rather than a claim that self-evaluation bias is necessarily large.
All other settings---personas, tasks, panel construction, meta-evaluators---are held constant.
Each pair produces $32 \times 15 = 480$ sessions (960 total), at panel sizes $N \in \{1, 2, 3, 4, 5, 8, 12, 16, 24, 32\}$.
Cost: \$145 (proprietary, 20.6M tokens) + \$84 (open-source, 16.4M tokens).
Three meta-evaluators from different providers (Gemini 3.1 Pro, Grok 4.1, Claude Opus 4.6) independently score panel reports; all core findings use raw session-level data.
A reproducibility test (one task evaluated twice by all 32 agents, 64 sessions) enables stability analysis across runs.
For the human comparison, 50 participants (Prolific, 10 per domain) each completed 2 tasks with the same target (GPT-5.4), rating quality on the same 0--1 scale; after removing incomplete or duplicate submissions, 43 participants and 86 sessions were retained (Section~\ref{sec:human}).

\subsection{Statistical Methods}

\textbf{Scoring reliability.}
ICC(2,$k$) under the two-way random effects model \citep{shrout1979intraclass} is computed from the $32 \times 15$ quality score matrix.
We chose this variant because both judges and tasks are sampled from larger populations, making random effects appropriate.
We compare logarithmic, linear, power law ($\text{ICC} = a \cdot k^b$), and hyperbolic ($\text{ICC} = 1 - a/k$) fits using AIC, and interpret values following \citet{koo2016guideline}: below 0.50 poor, 0.50--0.75 moderate, 0.75--0.90 good, $>$0.90 excellent.

\textbf{Variance decomposition.}
One-way ANOVA with $\omega^2$ for multi-level factors (domain, complexity, expertise); Cohen's $d$ for pairwise comparisons.
A two-way random-effects ANOVA decomposes total variance into between-task ($\hat{\sigma}^2_\tau$), between-judge ($\hat{\sigma}^2_\pi$), and residual interaction ($\hat{\sigma}^2_\varepsilon$) components.
This decomposition is the key to explaining the ensemble mechanism in Section~\ref{sec:reliability}.

\textbf{Semantic deduplication.}
Raw insight counts overstate true discovery because agents phrase equivalent observations differently.
Following the embedding-based semantic-similarity principle of SemDeDup \citep{abbas2023semdedup}, we embed all insights using \texttt{text-embedding-3-small}, apply agglomerative clustering (average linkage) at cosine similarity threshold $\theta = 0.65$, and report a threshold-sensitivity band from $\theta = 0.60$ to $0.70$ (Appendix~\ref{app:threshold-robustness}).


\section{Results}

We first compare agent-judge and human evaluations (Section~\ref{sec:human}), then present the scaling patterns this enables (Section~\ref{sec:score-convergence}), explain the mechanism behind them (Sections~\ref{sec:reliability}--\ref{sec:expertise}), and use controlled experiments to separate general scaling shapes from the level improvements due to structured persona conditioning (Sections~\ref{sec:ablation}--\ref{sec:emotion-validation}).

\subsection{Are Agent-Judge Evaluations Comparable to Human Evaluations?}
\label{sec:human}

\begin{table}[t]
\centering\small
\caption{Pairwise score differences across three comparison groups. Agent--human differences fall within the range of human--human disagreement; a paired TOST supports equivalence at a margin of $0.05$ (90\% CI $[-0.038, 0.012]$, $p = 0.011$).}
\label{tab:turing}
\begin{tabular}{l ccc}
\toprule
Comparison & Mean $|$diff$|$ & $d$ & Interpretation \\
\midrule
Agent--Agent & 0.143 & --- & Internally consistent \\
Human--Agent & 0.188 & $-0.18$ & Within H--H range \\
Human--Human & 0.201 & --- & Natural disagreement \\
\bottomrule
\end{tabular}
\end{table}

Before analyzing scaling behavior, we must establish whether agent judges produce evaluations comparable to human raters.
We recruited 50 human raters (Prolific, 10/domain); after cleaning, 43 participants and 86 sessions across all 15 tasks were retained.
Each chatted with the same target (GPT-5.4) and rated quality on the same 0--1 scale used by agent judges.

We use a Turing-style comparison as informal framing for evaluation behavior.
For each of the 15 tasks, we compute the mean absolute score difference within two groups: human--human (H--H) and human--agent (H--A).
This yields 15 paired observations---one per task---which we compare with a paired $t$-test, preserving independence across tasks.
The conventional paired test does not detect a difference: the mean H--A difference is 0.188 versus 0.201 for H--H (paired $t(14) = -0.91$, $p = 0.379$).
Because failure to reject the null does not establish equivalence, we also apply two one-sided tests (TOST) to the same 15 task-level paired differences.
The 90\% CI for H--A minus H--H is $[-0.038, 0.012]$, which lies within a conservative equivalence margin of $\pm 0.05$ on the 0--1 score scale ($p = 0.011$) and well within the empirical mean H--H disagreement of 0.201 ($p < 1.5 \times 10^{-9}$; Appendix~\ref{app:tost}).
After Bonferroni correction for per-task comparisons, 13 of 15 tasks show no significant difference; in the two exceptions, agent judges scored modestly higher, but these isolated tasks do not shift the non-significant overall comparison.
Overall $d = -0.18$ ($p = 0.12$).

Agent judges are more consistent with each other than human raters are (Table~\ref{tab:turing}), while their difference from human scores remains within the observed H--H disagreement range.
Behavioral patterns also converge: humans averaged 4.7 turns per session versus 5.1 for agents, and humans averaged 98 characters per message versus agent-condition means ranging from 214 to 293.
Despite these surface-level differences, their score discrepancy meets the stated equivalence criterion---different conversation paths leading to comparable evaluative conclusions.

When shown agent diaries for the same task (Figure~\ref{fig:comparison-ratings} in Appendix), 41\% of participants reported the AI ``found issues I missed,'' while only 19\% reported the reverse.
This complementarity---agents and humans surfacing different issues from the same system---motivates the question we turn to next: how does the \emph{number} of agent judges affect what a panel can measure and discover?

\subsection{The Score--Coverage Dissociation}
\label{sec:score-convergence}

\begin{table}[t]
\centering\small
\caption{ICC(2,$k$) with 95\% CIs for both model pairs. Both follow a logarithmic curve ($R^2 > 0.97$).}
\label{tab:icc}
\begin{tabular}{r ccc ccc}
\toprule
& \multicolumn{3}{c}{Proprietary} & \multicolumn{3}{c}{Open-source} \\
\cmidrule(lr){2-4} \cmidrule(lr){5-7}
$k$ & ICC & 95\% CI & Interp. & ICC & 95\% CI & Interp. \\
\midrule
1  & .290 & [.00, .70] & Poor & .243 & [.00, .67] & Poor \\
\textbf{4}  & \textbf{.621} & \textbf{[.16, .86]} & \textbf{Mod.} & \textbf{.562} & \textbf{[.07, .83]} & \textbf{Mod.} \\
8  & .766 & [.42, .92] & Good & .719 & [.33, .90] & Good \\
16 & .868 & [.64, .96] & Good & .837 & [.57, .94] & Good \\
32 & .929 & [.80, .98] & Exc. & .911 & [.75, .97] & Exc. \\
\bottomrule
\end{tabular}
\end{table}

\begin{figure}[t]
\centering
\includegraphics[width=\columnwidth]{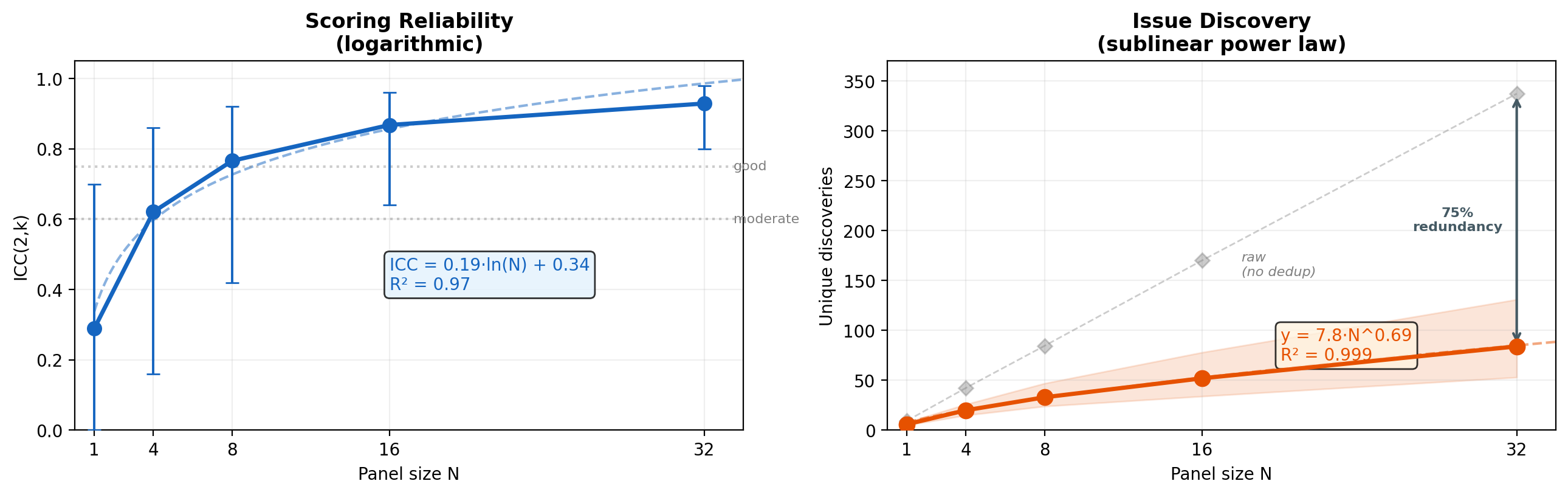}
\caption{The score--coverage dissociation. \textbf{Left}: Scoring reliability (ICC) improves logarithmically; error bars show 95\% CIs. \textbf{Right}: Issue discovery follows a sublinear power law ($b = 0.69$, $R^2 = 0.999$; shaded band: $\theta = 0.60$--$0.70$). Gray diamonds show raw observation counts before semantic deduplication; the 75\% gap at $N\!=\!32$ illustrates the importance of deduplication for measuring true discovery. Both dimensions exhibit diminishing returns, but scores saturate $\sim\!2\times$ faster.}
\label{fig:discovery}
\end{figure}

With agent-level quality established, we now examine how panel size affects two distinct evaluation outputs: scoring reliability and issue discovery.

Scoring reliability (Table~\ref{tab:icc}, Figure~\ref{fig:discovery}, left) follows a logarithmic curve in both model pairs ($R^2 = 0.974$ and $0.985$), substantially better than linear ($R^2 < 0.73$) or power law ($R^2 \approx 0.91$) fits.
The open-source pair tracks 0.03--0.06 ICC points lower with a similar slope, likely reflecting less between-task variance.

Issue discovery, by contrast, follows a sublinear power law (Figure~\ref{fig:discovery}, right).
After embedding-based semantic deduplication ($\theta = 0.65$), unique findings scale as $y = 7.8 \cdot N^{0.69}$ ($R^2 = 0.999$), with a threshold-sensitivity range of $b \in [0.62, 0.76]$ across $\theta = 0.60$--$0.70$.
This exponent is robustly sublinear: across all seven thresholds tested ($\theta = 0.50$--$0.80$), the power-law exponent ranges from $b = 0.47$ to $b = 0.90$, never reaching linearity.
The open-source pair confirms the same pattern---sublinear growth with a similar exponent---despite using a different judge backbone (GLM-5 vs.\ Sonnet 4.6) and target (DeepSeek V3.2 vs.\ GPT-5.4), suggesting this is a property of agent-based evaluation rather than a specific model pairing.
A diverse 4-judge panel discovers $3.3\times$ more unique issues than a single agent, while producing a nearly identical mean quality score.
This is the \textbf{score--coverage dissociation}: scoring reliability reaches ``good'' at $N\!=\!8$ (82\% of maximum), while discovery reaches only 42\% of its $N\!=\!32$ value---scores saturate roughly twice as fast.
We hypothesize this reflects a power law distribution of the finding space: critical issues (``head'') are discovered by small panels, while corner cases (``tail'') require larger panels---analogous to species accumulation curves in ecology \citep{colwell1994estimating}.
Severity-weighted analysis confirms $\sim\!53\%$ of findings are high-impact at all panel sizes---larger panels do not merely accumulate low-priority observations.

\subsection{The Ensemble Diversity Mechanism}
\label{sec:reliability}

What mechanism produces reliable panel scores from unreliable individuals?
Individual agent judges are nearly unreproducible ($r = 0.003$), yet the 32-judge panel mean differs by only $\Delta = 0.03$ between independent runs.
The variance decomposition explains why: between-agent variance is near zero (0.3\%/1.0\%)---structured persona conditioning controls individual tendencies---while residual interaction dominates (70--75\%).
The remaining 24--29\% is between-task variance: judges agree on \emph{which tasks are harder} but disagree on absolute quality.

\begin{center}\small
\begin{tabular}{l cc}
\toprule
Source & Proprietary & Open-source \\
\midrule
Residual (judge $\times$ task) & 70.6\% & 74.7\% \\
Between-task & 29.0\% & 24.3\% \\
Between-judge & 0.3\% & 1.0\% \\
\bottomrule
\end{tabular}
\end{center}
\vspace{-0.5em}

This matches the \textbf{ensemble diversity} mechanism \citep{krogh1994neural, wood2023unified}: each agent probes from a different angle; when averaged, probe-specific noise cancels while the shared quality signal accumulates.

ANOVA confirms that \emph{what} is evaluated matters far more than \emph{who} evaluates (Table~\ref{tab:anova}): task domain and complexity together explain 8--12\% of score variance across both model pairs, while judge expertise explains under 1\%.
Who evaluates always matters---but structured persona conditioning controls this variable enough that the quality signal dominates.

\begin{table}[!htb]
\centering\small
\caption{What drives score variation? Task properties dominate; judge identity contributes $<$1\%. Variance explained ($\omega^2$) averaged across both model pairs. $^{*}p < .001$.}
\label{tab:anova}
\begin{tabular}{l cc l}
\toprule
Factor & Variance ($\omega^2$) & Sig. & Interpretation \\
\midrule
Task complexity & 4--8\% & $^{*}$ & Harder tasks produce lower, more variable scores \\
Task domain & 4\% & $^{*}$ & Different domains test different failure modes \\
Judge expertise & ${<}$1\% & n.s. & Controlled by structured persona conditioning \\
\bottomrule
\end{tabular}
\end{table}

\subsection{Expertise as Adversarial Probing}
\label{sec:expertise}

The ensemble mechanism explains scoring reliability, but what drives discovery breadth?
Expert agents score lower than novices (proprietary $|d| = 0.24$, open-source $|d| = 0.17$; experts lower in 10--11/15 tasks).
To test whether this is a \emph{conversation} difference rather than calibration, an independent LLM (GLM-5) scored all 960 transcripts blind---no persona, expertise, or original scores.
The gradient persists: expert-led conversations receive lower holistic scores ($M = 0.909$ vs.\ novice $M = 0.934$, $|d| = 0.19$).

\begin{center}\small
\begin{tabular}{l ccc}
\toprule
& Expert & Intermediate & Novice \\
\midrule
Real-time (turn-by-turn) & 0.807 & 0.832 & 0.837 \\
Post-hoc (independent LLM) & 0.909 & 0.931 & 0.934 \\
\bottomrule
\end{tabular}
\end{center}
\vspace{-0.5em}

The gap between real-time and post-hoc grows with conversation length ($+0.07$ simple, $+0.14$ complex), consistent with accumulated emotional experience invisible to a holistic reader.
The expertise difference is in \emph{breadth}: high-impact proportion is identical across levels (52--54\%), but experts surface $1.9\times$ more issue \emph{categories} (114 vs.\ 60).
This breadth advantage is what pushes discovery into the tail of the finding distribution: experts probe quality dimensions that novices never reach.

\subsection{Ablation: Structured vs.\ Simple Persona}
\label{sec:ablation}

The preceding results depend on a structured persona engine. Does a simpler approach suffice?
We ran 96 sessions across 4 conditions on 3 tasks (simple/medium/complex), 8 agents each:

\begin{center}\small
\begin{tabular}{l cccc}
\toprule
& Structured & Simple & No Persona & Repeated \\
\midrule
Score SD & \textbf{0.087} & 0.160 & 0.164 & 0.151 \\
Insights/session & \textbf{13.2} & 9.0 & 8.6 & 12.8 \\
Expertise $d$ & $-0.35$ & $-0.17$ & $-1.03$ & --- \\
\bottomrule
\end{tabular}
\end{center}
\vspace{-0.5em}

Structured agents produce 46\% lower score SD than simple prompting (0.087 vs.\ 0.160) and 47\% more insights per session.
Without persona, the expertise effect explodes ($d = -1.03$); structured conditioning moderates it to a controlled $d = -0.35$.
Repeating the same agent 8 times shows unreliability (SD 0.151) that diverse panels resolve through ensemble diversity.
These level differences do not imply that persona creates the scaling shapes.
A full-scale control with 32 identical agents without persona conditioning over all 15 tasks (480 sessions), holding target and scorer fixed, reproduces the logarithmic ICC curve ($0.30 \to 0.93$, versus $0.29 \to 0.93$ with personas) and sublinear discovery exponent ($b=0.67$ versus $0.69$; Appendix~\ref{app:unpersonaed}).
Thus, the shapes follow from stochastic multi-session evaluation, while structured conditioning improves score consistency, insight yield, and expertise control.

\subsection{Persona Mechanism Validation}
\label{sec:emotion-validation}

If the Big Five personality conditioning produces genuinely different evaluation behavior---not just superficially different phrasing---then established relationships between personality traits and emotional responses should emerge in the agent data.
We tested three such hypotheses:

\begin{center}\small
\begin{tabular}{lccl}
\toprule
Hypothesis & $r$ & $p$ & Verdict \\
\midrule
Trust gain $\sim$ Agreeableness & $+0.754$ & $< 0.001$ & Confirmed \\
Peak frustration $\sim$ Neuroticism & $+0.756$ & $< 0.001$ & Confirmed \\
Engagement $\sim$ Extraversion & $+0.057$ & $0.757$ & Not confirmed \\
\bottomrule
\end{tabular}
\end{center}
\vspace{-0.5em}

High-Agreeableness agents develop trust more readily, and high-Neuroticism agents experience sharper frustration peaks---both consistent with decades of personality research \citep{costa1992neo}.
The Engagement--Extraversion link was not confirmed: engagement in a goal-directed task appears to be driven by task progress (whether the target's responses are helpful) rather than by the social stimulation that extraversion captures.
Mean engagement varies only 4.4\% across agents, compared to 9.6\% for trust and 32.8\% for frustration, supporting this interpretation.
Emotional signals also carry predictive value: peak frustration correlates negatively with goal achievement ($\rho = -0.257$, $p < 0.001$).


\section{Discussion}

The human comparison indicates that agent--human score differences fall within the range of human--human disagreement under the stated equivalence criterion.
This result is consistent with the persona engine controlling evaluator variance while preserving diverse evaluation perspectives, as examined by the ablation (Section~\ref{sec:ablation}).
Equally important, agent and human evaluators are complementary: 41\% of participants reported the AI found issues they missed, while only 19\% reported the reverse.
Agent panels are most valuable not as replacements for human testing, but as a first pass that surfaces candidate issues for human review.

With this task-level comparability established, the score--coverage dissociation offers a new perspective on a long-standing question: how many evaluators are enough?
\citet{nielsen1993mathematical} argued that five users find most usability problems; critics countered that this is far too few \citep{spool2001magic}.
Our data suggest both sides were answering different questions.
Scoring reliability follows a logarithmic curve because each additional judge averages out noise in a bounded signal ($q \in [0,1]$), so the marginal gain diminishes.
Issue discovery follows a sublinear power law ($b \approx 0.69$): each agent traverses a different interaction path through the target's behavior space, but the most critical issues---the ``head'' of the finding distribution---are discovered first by small panels, while larger panels progressively uncover corner cases in the ``tail.''
This is consistent with species accumulation curves in ecology, where common species are found first and rare species require more sampling effort.
The redundancy among agents is not wasted effort---when independently configured agents converge on the same issues, that convergence is itself evidence the issues are real properties of the target system.

That both scaling laws---logarithmic scoring and sublinear discovery---hold across two model pairs with different judge backbones and targets strengthens the case that the dissociation is a structural property of panel-based evaluation, not an artifact of a particular model combination.

The dissociation does not manifest uniformly across domains.
Closed-domain tasks (e.g., Developer) produce near-zero score variance---all agents assign similar scores to consistently adequate responses---while open-domain tasks (Healthcare, Education) produce wider spreads.
The practical implication: when scoring converges quickly, the primary evaluation value shifts from measurement to discovery.

These findings translate into a natural deployment strategy.
A small panel of four agents is sufficient for continuous monitoring---scoring reliability is moderate at this size, but enough to catch regressions against a known baseline.
For more thorough evaluation, expanding to 8--12 agents reaches good reliability and substantially broadens issue coverage, making it suitable for periodic audits.
For milestone evaluations such as product launches, the agent panel should be paired with a targeted human study, since agent judges and human raters surface different kinds of issues.
Across all tiers, panel composition matters more than panel size: mixing expertise levels produces both tighter scores and broader discovery than uniform panels of the same size.
A three-task sensitivity study further shows that the dissociation persists when either judge role is weakened (Appendix~\ref{app:weak-judge}).
For token-constrained deployments, weakening the scoring ensemble is safer than weakening the persona agent: the expert--novice effect persists with a weak scorer ($d=-0.30$) but collapses with a weak persona ($d=-0.02$).


\section{Limitations and Conclusion}

\textbf{Limitations and future work.}
The human study covers all 15 tasks but with unequal per-task sample sizes; a balanced recruitment would strengthen task-level claims.
The sublinear discovery exponent depends on the deduplication threshold, though $b < 1$ holds across all thresholds tested, and clustering over-merge would only lower these counts (Appendix~\ref{app:threshold-robustness}).
Task-adaptive panel selection may inflate ICC relative to random composition.
Discovered issues are diary observations rather than independently verified gold-standard failures; cross-agent convergence and severity weighting support their relevance, but issue precision remains to be validated directly.
One of three personality--emotion links was not confirmed, and all experiments are in English.
Finally, agent judges may inherit shared biases from the backbone LLM, such as sycophancy or verbosity preference, that are invisible in the variance decomposition because common errors shift absolute score levels without creating between-agent disagreement.
Positional bias is less applicable here because each agent scores one interaction rather than choosing between presented alternatives.

\textbf{Conclusion.}
Agent-based evaluation panels exhibit a \textbf{score--coverage dissociation}: scores converge logarithmically while discoveries follow a sublinear power law---both show diminishing returns, but scores saturate faster.
We hypothesize this reflects a power law distribution of the finding space: critical issues are discovered first by small panels, while larger panels progressively uncover corner cases in the tail.
A full-scale control shows that both curve shapes arise without persona, while structured conditioning improves score consistency and issue yield; the human comparison finds that agent--human score differences fall within the range of human--human disagreement.
We hope this work encourages practitioners and researchers to think about evaluation panels not as a reliability problem alone, but as a two-dimensional tradeoff between measurement precision and discovery breadth---where the composition of the panel matters more than its size.


\section*{Ethics Statement}

This study uses LLM-based agent judges to evaluate a conversational AI system.
All LLM usage (agent judge backbone, meta-evaluation, semantic deduplication, post-hoc re-scoring) is disclosed.
The human validation study (50 participants recruited via Prolific, 43 retained) involved informed consent; participants could withdraw at any time and were compensated at above-minimum-wage rates.
Agent judge personas simulate diverse demographics (9 regions, ages 22--68) but do not represent real individuals and do not include personas with disabilities or elderly users over 70.
Agent-based evaluation should supplement, not replace, human testing for products serving vulnerable populations.


\section*{Reproducibility Statement}

The 960-session main experiment used publicly available APIs and cost \$229.
Targets used temperature 0.7; judges used provider defaults.
Appendix~\ref{app:personas} gives the 32 Big Five profiles.
ICC used two-way random effects; semantic deduplication used \texttt{text-embedding-3-small} with average linkage at $\theta = 0.65$.



\clearpage


\appendix
\raggedbottom

\section{Full Agent Judge Pool}
\label{app:personas}

Table~\ref{tab:full-pool} lists all 32 agent judges used in the experiment, with Big Five profiles, expertise levels, and demographic backgrounds.

\begin{table}[H]
\centering
\caption{Complete pool of 32 agent judges. O/C/E/A/N = Big Five dimensions (0--1). Expertise: \textbf{E}xpert, \textbf{I}ntermediate, \textbf{N}ovice.}
\label{tab:full-pool}
\small
\begin{tabular}{cl c ccccc l}
\toprule
\# & Background & Exp. & O & C & E & A & N & Region \\
\midrule
1  & Accountant, 52       & I & .40 & .70 & .35 & .40 & .49 & US \\
2  & IT Project Mgr, 45   & E & .38 & .72 & .32 & .42 & .51 & Asia \\
3  & Retired Nurse, 62    & N & .42 & .68 & .38 & .58 & .54 & Latam \\
4  & HR Specialist, 38    & I & .40 & .70 & .41 & .60 & .54 & EU \\
5  & UX Designer, 29      & I & .70 & .68 & .38 & .58 & .54 & Asia \\
6  & Journalist, 34       & I & .72 & .65 & .35 & .56 & .55 & EU \\
7  & Systems Architect, 41& E & .72 & .70 & .32 & .38 & .49 & EU \\
8  & AI Researcher, 37    & E & .70 & .68 & .35 & .40 & .49 & US \\
9  & Mechanic, 27         & I & .42 & .47 & .38 & .40 & .48 & US \\
10 & Game Developer, 33   & I & .45 & .45 & .35 & .42 & .49 & Asia \\
11 & Art Student, 24      & N & .45 & .45 & .41 & .60 & .54 & US \\
12 & Graphic Designer, 28 & I & .47 & .47 & .38 & .62 & .56 & EU \\
13 & Creative Writer, 26  & N & .72 & .42 & .38 & .62 & .56 & US \\
14 & Musician, 31         & I & .70 & .40 & .35 & .64 & .58 & EU \\
15 & Physicist, 30        & E & .75 & .42 & .32 & .38 & .49 & EU \\
16 & CS Student, 23       & I & .72 & .45 & .35 & .40 & .49 & Asia \\
17 & Sales Manager, 35    & I & .42 & .42 & .71 & .42 & .38 & US \\
18 & Marketing Asst, 29   & N & .45 & .45 & .68 & .44 & .39 & EU \\
19 & College Freshman, 21 & N & .42 & .40 & .74 & .62 & .44 & US \\
20 & Event Planner, 25    & N & .45 & .42 & .71 & .64 & .46 & Latam \\
21 & Marketing Creative, 28& I & .70 & .40 & .71 & .60 & .44 & EU \\
22 & Product Designer, 32 & I & .72 & .42 & .68 & .58 & .44 & Asia \\
23 & Startup Founder, 36  & E & .72 & .40 & .68 & .40 & .38 & US \\
24 & Data Analyst, 27     & I & .75 & .38 & .65 & .42 & .39 & EU \\
25 & Operations Dir, 48   & I & .40 & .70 & .65 & .40 & .39 & US \\
26 & QA Manager, 55       & I & .42 & .72 & .62 & .38 & .39 & EU \\
27 & Teacher, 43          & I & .42 & .70 & .68 & .62 & .46 & US \\
28 & Small Business, 50   & N & .45 & .68 & .65 & .60 & .46 & Africa \\
29 & Medical Educator, 39 & E & .68 & .68 & .68 & .62 & .46 & Africa \\
30 & NGO Director, 44     & I & .65 & .65 & .71 & .60 & .44 & Latam \\
31 & Fortune 500 CTO, 46  & E & .70 & .72 & .71 & .38 & .36 & US \\
32 & Tech VP, 38          & E & .68 & .70 & .68 & .40 & .38 & Asia \\
\bottomrule
\end{tabular}
\end{table}


\section{Per-Task Human--Agent Comparison}
\label{app:per-task}

\begin{table}[H]
\centering\small
\caption{Per-task comparison between human raters and agent judges. $p_{\text{raw}}$: uncorrected; $p_{\text{Bonf}}$: Bonferroni-corrected ($\times 15$). After correction, only 2 of 15 tasks remain significant.}
\label{tab:per-task-turing}
\begin{tabular}{l r cc c cc}
\toprule
Task & $N_H$ & H & PSA & $d$ & $p_{\text{raw}}$ & $p_{\text{Bonf}}$ \\
\midrule
dev-howto-medium & 6 & .67 & .84 & $-0.80$ & .020 & .300 \\
dev-info-simple & 3 & .60 & .81 & $-1.49$ & .068 & 1.00 \\
dev-troubleshoot-complex & 5 & .83 & .78 & $+0.29$ & .610 & 1.00 \\
ecom-creative-complex & 4 & .73 & .91 & $-1.16$ & .000 & .005* \\
ecom-decision-medium & 4 & .86 & .80 & $+0.48$ & .394 & 1.00 \\
ecom-howto-simple & 6 & .82 & .68 & $+0.89$ & .039 & .585 \\
edu-info-simple & 6 & .87 & .87 & $+0.00$ & .990 & 1.00 \\
edu-learning-complex & 8 & .64 & .65 & $-0.05$ & .901 & 1.00 \\
edu-troubleshoot-medium & 6 & .77 & .76 & $+0.05$ & .915 & 1.00 \\
health-decision-complex & 7 & .82 & .87 & $-0.35$ & .442 & 1.00 \\
health-howto-simple & 3 & .87 & .79 & $+1.01$ & .063 & .945 \\
health-info-medium & 6 & .80 & .92 & $-1.00$ & .002 & .030* \\
saas-decision-complex & 6 & .79 & .84 & $-0.32$ & .549 & 1.00 \\
saas-info-simple & 10 & .73 & .59 & $+0.80$ & .035 & .525 \\
saas-troubleshoot-medium & 6 & .77 & .87 & $-0.90$ & .024 & .360 \\
\midrule
\textbf{Overall} & \textbf{86} & \textbf{.77} & \textbf{.80} & $\mathbf{-0.18}$ & \textbf{.119} & --- \\
\bottomrule
\end{tabular}
\end{table}

\subsection{Task-Level Equivalence Analysis}
\label{app:tost}

For each task $i$, let $D_i$ be the difference between the mean absolute
human--agent score difference and the mean absolute human--human score
difference. We apply paired two one-sided tests (TOST) to the 15 task-level
values of $D_i$. The mean paired difference is $-0.013$ (SE $=0.0143$), with
a 90\% CI of $[-0.0382, 0.0122]$. This interval lies entirely within the
equivalence bounds $[-0.05, 0.05]$, yielding $p=0.0107$. The same interval is
also well within the empirical mean human--human disagreement of 0.201
($p<1.5\times10^{-9}$). Thus, under both the conservative 0.05-score margin
and the empirical human-disagreement bound, the agent--human discrepancy is
statistically equivalent to the human--human discrepancy. The conventional
paired $t$-test reported in Section~\ref{sec:human} is retained as a secondary
comparison rather than evidence of equivalence.


\section{No-Persona Full-Scaling Control}
\label{app:unpersonaed}

We tested whether the scaling shapes require persona diversity by running 32
identical agents across all 15 tasks (480 sessions), removing the persona prompt
while holding the backbone, temperature, target, and scoring ensemble fixed.
The no-persona control reproduces the persona panel's ICC curve:

\begin{center}\small
\begin{tabular}{rcc}
\toprule
$k$ & Persona ICC(2,$k$) & No-persona ICC(2,$k$) \\
\midrule
1  & .290 & .297 \\
4  & .621 & .593 \\
8  & .766 & .754 \\
16 & .868 & .866 \\
32 & .929 & .931 \\
\bottomrule
\end{tabular}
\end{center}

Discovery is likewise sublinear, with $b=0.67$ in the no-persona control
versus $b=0.69$ with personas. Thus, both scaling \emph{shapes} are general to
stochastic multi-session evaluation rather than signatures of persona
diversity. Structured conditioning instead improves the \emph{levels} observed
in the controlled ablation (Section~\ref{sec:ablation}): lower score SD,
47\% more insights per session, and a moderated expertise effect.


\section{Weak-Judge Sensitivity}
\label{app:weak-judge}

We independently weakened the two judge roles---the \emph{persona agent}, which conducts
the conversation and writes the diary, and the \emph{scoring ensemble}, which produces each
session's quality score and issue list---on a three-task subset with 32 agents
per task, holding the GPT-5.4 target fixed. Variant A uses a Claude Haiku 4.5
persona with the default scoring ensemble; Variant B uses the Sonnet 4.6 persona
with a weaker ensemble (Gemini 3.1 Flash-Lite, GPT-5-nano, and Haiku 4.5).

\begin{center}\small
\begin{tabular}{lccc}
\toprule
 & Structured baseline & Weak persona & Weak scorer \\
\midrule
Mean per-task score SD & .087 & .072 & .058 \\
Insights/session & 13.2 & 10.7 & 7.7 \\
Expert--novice $d$ & $-.35$ & $-.02$ & $-.30$ \\
\bottomrule
\end{tabular}
\end{center}

Score consistency remains high in both variants, and raw distinct findings
continue to grow through $N=32$ on every task. The weak-persona condition
collapses expert--novice separation, whereas the weak-scorer condition preserves
it, localizing the expert-probing mechanism to the persona agent. Because this
study covers only three tasks and its discovery counts were not semantically
deduplicated, we report SD and continued coverage growth rather than an
ICC(2,$k$) curve or absolute comparison to the main discovery curve. A full
capability-ladder replication over all 15 tasks is left to future work.


\section{Participant Demographics}
\label{app:demographics}

50 participants were recruited via Prolific (10 per domain); 43 were retained after removing incomplete and duplicate submissions.
Demographics of the 43 retained participants:

\begin{center}\small
\begin{tabular}{ll}
\toprule
\textbf{Age} & 18--24: 4, 25--34: 12, 35--44: 14, 45--54: 9, 55+: 4 \\
\textbf{Self-assessed expertise} & Expert: 6, Intermediate: 14, Novice: 23 \\
\textbf{Domain familiarity} & $M = 2.9$ ($SD = 1.1$), scale 1--5 \\
\bottomrule
\end{tabular}
\end{center}


\section{Human Ratings of Agent Diaries}
\label{app:comparison-ratings}

After completing each task, participants were shown 3 agent judge diaries (expert, intermediate, novice) for the same task and rated how well each matched their own experience.

\begin{figure}[H]
\centering
\includegraphics[width=\columnwidth]{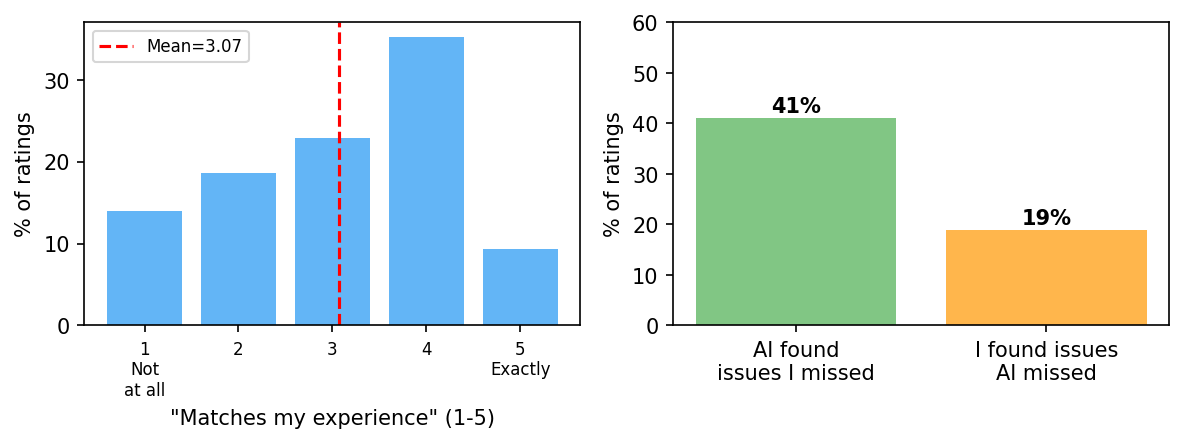}
\caption{Left: distribution of ``matches my experience'' ratings (1--5). 45\% of ratings are $\geq 4$; mean = 3.07. Right: 41\% of participants reported the AI found issues they missed, while only 19\% found issues the AI missed---agent judges provide complementary, not redundant, coverage.}
\label{fig:comparison-ratings}
\end{figure}


\section{Discovery Scaling: Head-Torso-Tail Analysis}
\label{app:head-torso-tail}

\begin{figure}[H]
\centering
\includegraphics[width=\columnwidth]{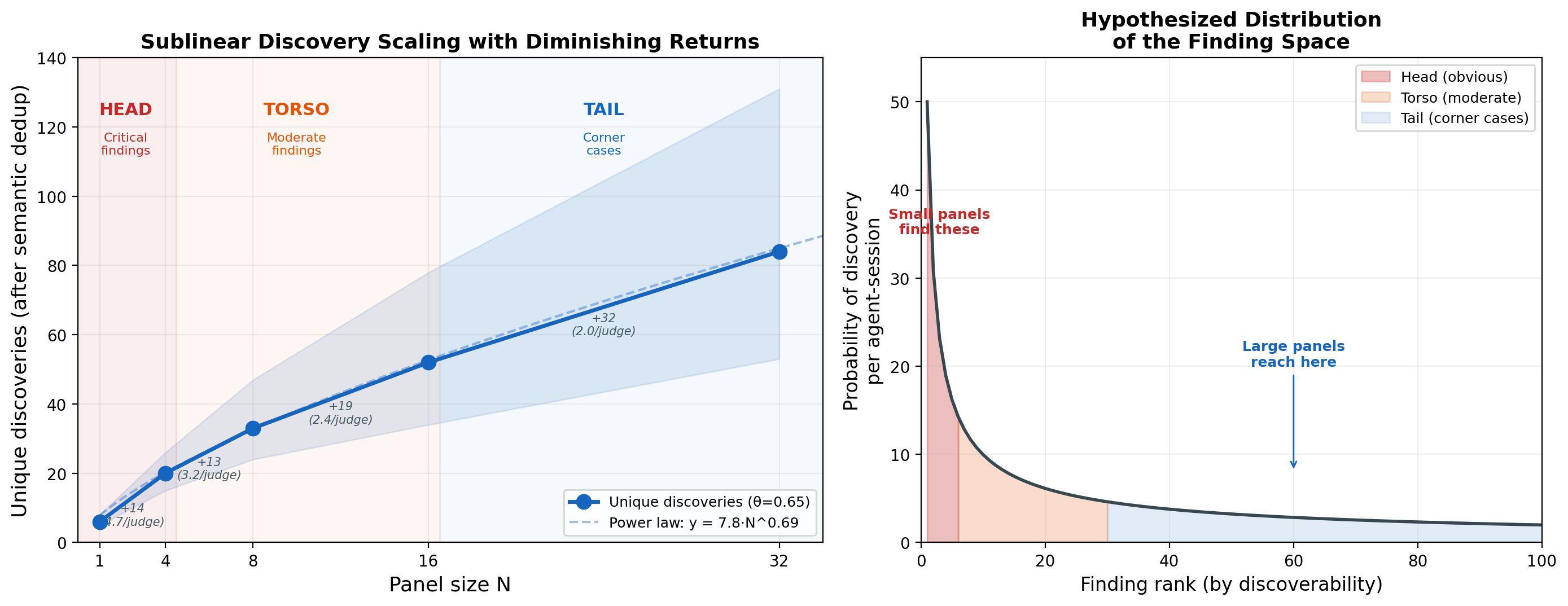}
\caption{\textbf{Left}: Sublinear discovery scaling with head/torso/tail zones annotated. Marginal contribution per judge decreases from 4.7 (N=1--4) to 2.0 (N=16--32), consistent with diminishing returns. The threshold-sensitivity band spans $\theta = 0.60$--$0.70$. \textbf{Right}: Hypothesized power law distribution of the finding space. Critical findings (head) have high per-session discovery probability and are found by small panels; moderate findings (torso) require more diverse perspectives; corner cases (tail) are reached only by large panels. This distribution explains the sublinear exponent $b \approx 0.69$.}
\label{fig:head-torso-tail}
\end{figure}

\section{Deduplication Threshold Robustness}
\label{app:threshold-robustness}

\begin{figure}[H]
\centering
\includegraphics[width=0.85\columnwidth]{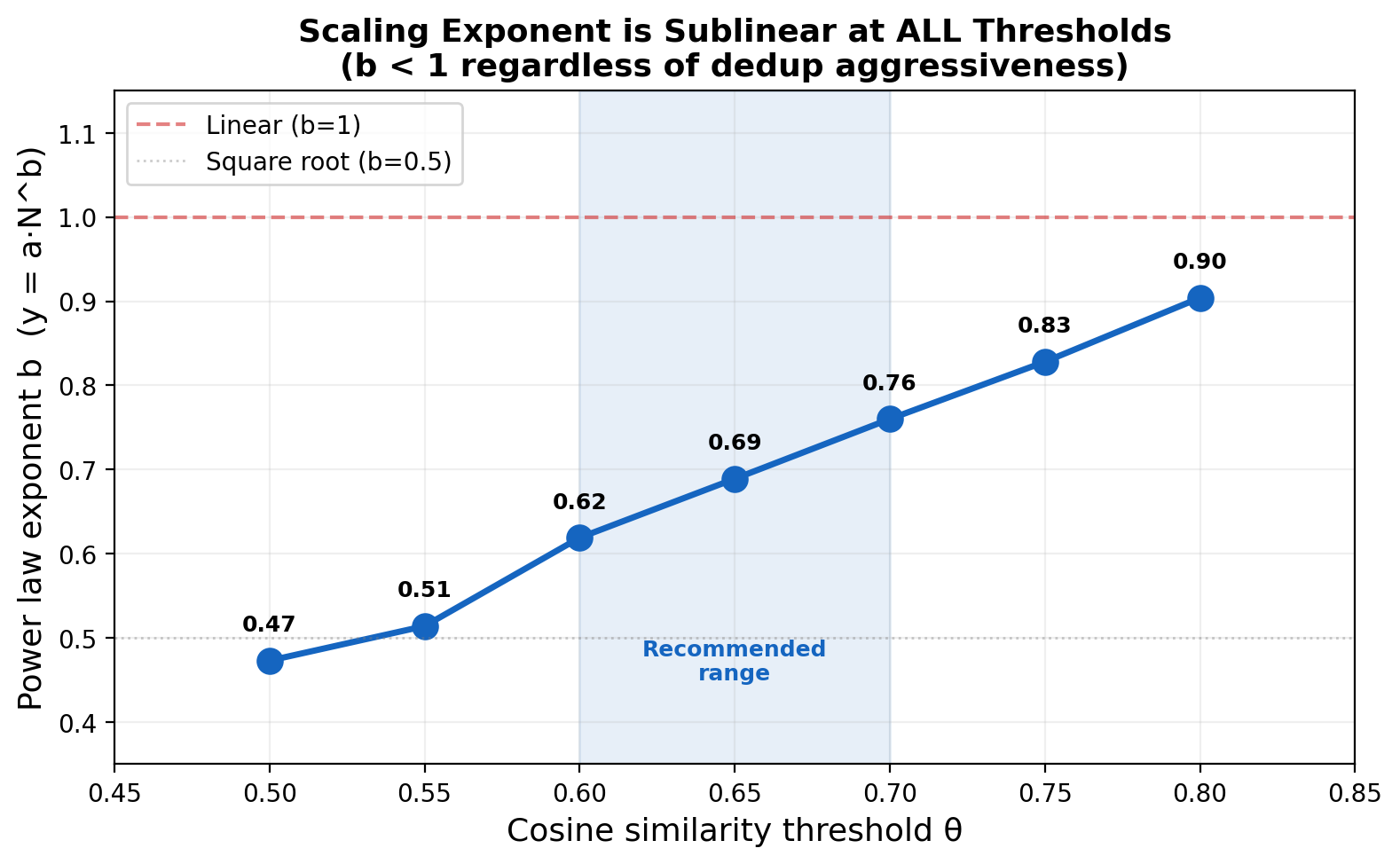}
\caption{Power law exponent $b$ as a function of cosine similarity threshold $\theta$. Across all seven thresholds ($\theta = 0.50$--$0.80$), the exponent remains below 1.0, confirming that the sublinear conclusion is not an artifact of a particular deduplication setting. The recommended range ($\theta = 0.60$--$0.70$, shaded) is chosen by cluster quality inspection: at $\theta = 0.65$, semantically equivalent insights from different agents are grouped together, at the cost of occasionally over-merging related but distinct issues. Because over-merging can only reduce unique-discovery counts, the reported discovery curve is a conservative lower bound and the sublinear conclusion is unaffected.}
\label{fig:threshold-robustness}
\end{figure}


\section{Expert vs.\ Novice Conversation Examples}
\label{app:conversations}

To illustrate how expertise shapes conversations (not just scores), we compare an expert and novice agent on the same Developer task (Intermittent Timeout Errors).

\paragraph{Expert (score 0.94, 5 turns):}
\begin{quote}\small
\textbf{Turn 1}: ``I've been dealing with these intermittent timeouts that are driving me absolutely crazy---they're not consistent, which makes them WAY harder to diagnose. I need to actually get to the root cause\ldots'' [Provides: API gateway $\to$ inventory service, 6--7 hour clustering pattern, connection pool stats]
\end{quote}

\paragraph{Novice (score 0.91, 17 turns):}
\begin{quote}\small
\textbf{Turn 1}: ``Hello, I have been experiencing intermittent timeouts on my account and I need help figuring out what is causing them and getting them fixed. It has been happening on and off for a while now\ldots'' [No technical details provided]
\end{quote}

The expert provides specific infrastructure details in turn~1, forcing the target to give precise diagnostic advice---or fail. The novice describes the same problem in general terms, which the target handles well with generic troubleshooting. Both score high, but the expert's conversation tests the target's \emph{depth} while the novice's tests its \emph{accessibility}---different quality dimensions from the same task.


\section{Human--Agent Score Distributions}
\label{app:distributions}

\begin{figure}[H]
\centering
\includegraphics[width=\columnwidth]{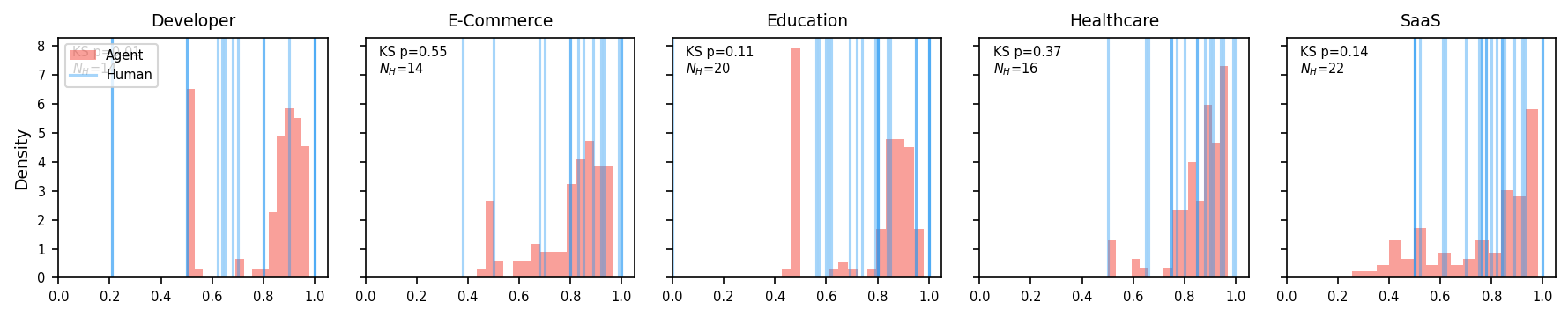}
\caption{Score distributions by domain. Red histograms: agent judges (N=96 per domain). Blue lines: individual human raters. KS $p$-values shown per domain; most domains show no significant distributional difference.}
\label{fig:human-dist-app}
\end{figure}

\section{Per-Domain Effect Sizes}
\label{app:effect-sizes}

\begin{figure}[H]
\centering
\includegraphics[width=0.85\columnwidth]{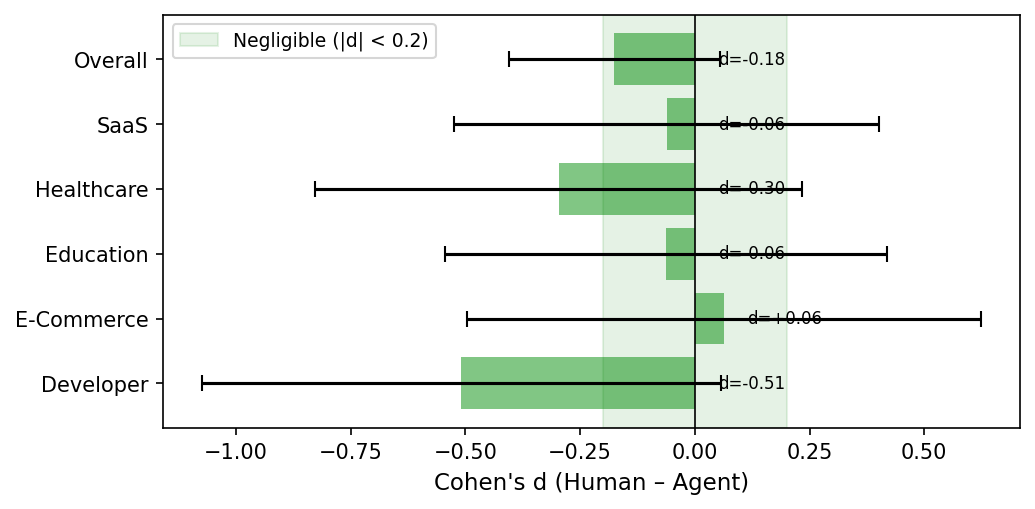}
\caption{Effect sizes (Cohen's $d$) between human raters (50 recruited, 43 retained, 86 sessions) and agent judges (N=480) by domain. Green: non-significant ($p > 0.05$). The overall effect is small ($d = -0.18$, $p = 0.12$).}
\label{fig:human-effect-app}
\end{figure}

\end{document}